# FAF: A novel multimodal emotion recognition approach integrating face, body and text

Zhongyu Fang[1], Aoyun He[1], Qihui Yu[1], Baopeng Gao[1], Weiping Ding[1]*, Tong Zhang[2]*, Lei Ma[1]*

1. School of Information Science and Technology, University of Nantong; Nantong, China, CN 226019
2. School of Computer science, Nanjing University of Science and Technology: Nanjing, China, CN 250000*

*Correspondence: malei@ntu.edu.cn; ding.wp@ntu.edu.cn; tong.zhang@njust.edu.cn;

**Abstract:**

Multimodal emotion analysis performed better in emotion recognition depending on more comprehensive emotional clues and multimodal emotion dataset. In this paper, we developed a large multimodal emotion dataset, named "HED" dataset, to facilitate the emotion recognition task, and accordingly propose a multimodal emotion recognition method. Specifically, the "HED" dataset contains happy, sad, disgust, angry and scared emotion-aligned face, body and text samples, which are much larger than existing datasets. Moreover, the emotion labels were correspondingly attached to those samples by strictly following a standard psychological paradigm. To promote recognition accuracy, "Feature After Feature" framework was used to explore crucial emotional information from the aligned face-body-text samples. For the images, a residual network was used for feature extraction. To understand the text, we use the BERT word vector for feature selection. To fuse all the emotion clues, the image features were initially fused, and the fused feature vectors were stitched with text features to form the combined features. Then, the combined features are further explored by using convolutional layers to explore the high-level complementary information among the multimodal information, and the attention mechanism was introduced to give different weights and improve the performance of emotion recognition in the fused modality. We employ various benchmarks to evaluate the "HED" dataset and compare the performance with our method. The results show that the five-classification accuracy of the proposed multimodal fusion method is about 83.75%, and the performance is improved by 1.83%, 9.38%, and 21.62% respectively compared with that of individual modalities. The complementarity between each channel is effectively used to improve the performance of emotion recognition. We had also established a multimodal online emotion prediction platform, aiming to provide free emotion prediction to more users.

## 1. Introduction

Emotion recognition is an important area of psychological and computer research. How to improve the accuracy of emotion recognition has become a primary issue. In recent years, with the continuous development of artificial intelligence technology, human-computer interaction has become the focus of research in the field of information science. As one of the critical technologies to realize human-computer interaction, emotion recognition has gradually received a lot of attention from researchers. At present, most of the research works on emotion recognition are based on single-modal, such as facial expressions [1-3], body movements [4-5] and speech text [6-7]. However, emotion recognition based on unimodal often has limitations and, in most cases, could only reflect a portion of human emotional expression. Multimodal emotion recognition can link individual unimodal channels and use the feature complementarity between channels to combine multiple information to determine the emotional state. Studies have shown that the multimodal emotion recognition approach has better performance than unimodal emotion judgment in most cases [8].

The difficulty of multimodal recognition is not only to control the internal information of individual modality (Intra-modality), but also to complement the interactive features between individual modalities (Inter-modality). It has been extensively studied by scholars, such as Tensor Fusion Network (TFN) proposed by Zadeh et al [9], Polynomial Tensor Pooling (PTP) proposed by Hou et al [10], and Memory Fusion Network (MFN) presented by Zadeh et al [11]. (MFN) proposed by Zadeh et al [11], etc. These models have achieved good results by adding interactions among the modalities based on the use of information within a single modality. However, considering the time complexity and space complexity as well as the redundancy of high-dimensional feature information, especially in the face of problems such as the excessive dimensionality of fused features, the above methods need to be further explored in multimodal fusion feature processing. In the emotion recognition field, the current multimodal field tends to combine human facial expressions with modalities such as voice and text, and there are also studies that combine expressions with physiological factors such as EEG and ECG, which significantly improve the emotion recognition results. However, some studies have shown that body posture is also an important way to represent human emotions [12]. Also, the quality of dataset is one of the key factors affecting deep learning. The existing datasets such as Extended Cohn-Kanada (CK+) collected by P. Lucy, CK+ dataset contains 327 sequences of labeled expression pictures of 123 objects, which are divided into seven expressions of normal, angry, contempt, disgust, fear, happy and sad, the number is small and not conducive to large-scale training; EmotionNet dataset collected from Internet about one million images, which contains basic expressions, compound expressions, and the annotation of expression units, with a slightly lower correct rate; Fer2013 contains a total of 26190 48*48 grayscale images with six kinds of expressions, with a lower resolution.

In summary, our work make the following contributions: Firstly, we build and expose sentiment datasets based on facial emotions and body gestures, which were collected from the Internet and manually filtered and detected; Secondly, we propose a novel multimodal sentiment recognition method that integrates face, body and text; Finally, we build a multimodal sentiment detection platform based on the FAF framework approach to provide free sentiment prediction to more users

This paper is organized as follows: Section 2 introduces the work related to multimodal emotion recognition; Section 3 details the specific implementation of multimodal fusion; Section 4 presents the experimental results data and the analysis of the results; the last section concludes the paper and looks forward to the next work.

## 2. Related work

With the development of deep learning techniques, some of the complex machine learning problems of the classification have been solved. Deep learning has greatly contributed to the solution of larger and more complex multimodal analysis problems [15]. Neural networks based multimodal emotion recognition problems has gradually become a research direction for many scholars.

Modal fusion approaches can be broadly classified into four types, which are data-level fusion, feature-level fusion, decision-level fusion, and model-level fusion [16]. For example, Minotto et al [17] used SVM for sensing-level fusion to further mine the modal data information; Wang et al [18] presented SAL (Select-Additive Learning) to stitch multiple modal features before correlation, and achieved better results; Liu et al [19] proposed a multi-level fusion method, which effectively enriched the semantics of the upper and lower layers; Middya et al [20] proposed model layer fusion based on audio data, which improved the generalization ability of the model in multimodal sentiment analysis. In terms of emotion recognition, Jiang et al [21] combined facial expressions and body movements to more effectively tap the deep semantic correspondence between them; Morency et al [22] expanded multimodal emotion recognition to three types of information: image, text and speech, and further complemented the multimodal dataset; Nguyen et al [23] even presented to combine facial expressions, pose body movements and voice, among other sources, were combined to recognize emotions. Unlike physical information, research scholars have further explored the physiological information evoked by emotions. Zhang et al [24] combined physiological information such as EEG and ECG and mapped them at a high level to effectively improve the accuracy of emotion recognition; Katsigiannis et al [25] provided a multimodal database of physiological information to further promote the development of emotion recognition research.

As discussed, although many efforts have been made on multimodal sentiment recognition methods, the mentioned algorithms have the following limitations and challenges. (1) The quality and quantity of previous datasets need to be extended. (2) The multimodal fusion algorithm is not accurate enough for the extraction of combined features after feature fusion. (3) Time complexity and space complexity are not ideal and waste excessive computational resources in some aspects. Inspired by the above work, this paper proposes a multimodal emotion recognition method based on face, gesture and text. Unlike the early fusion methods, it has further research in the processing of fused features after fusion. Considering combining the internal information in a single modality and the interaction information between modalities and solving the problems of information redundancy and high dimensionality, this paper achieves the best result of tri-modal fusion by combining features followed by a two-dimensional convolution operation and adding attention weights. To address the problems of insufficient data volume and excessive irrelevant noise in existing datasets, this paper openly constructs its own facial emotion and body movement datasets for sentiment analysis research.

## 3. Dataset

4. Tree structure was used to build the dataset, with five emotion nouns as subtrees (angry, disgust, happy, sad, and sacred), each including three modalities (face, body, and text. where the face and body) were segmented from the images as face and body, and the background and associated text of the images were extracted as text. The image dataset was acquired by two experts in psychology from various sources such as movies and TV shows. Among them, the ages include young children, young adults, and old adults; the geographical areas include Asia, Africa, and Europe. In this dataset, five of these emotion categories are taken and the dataset has a total sample size of 17,441 after data enhancement. The samples of facial and physical emotion categories are shown in Figure 1 and Figure 2. The number of each sample is shown in Table 1.

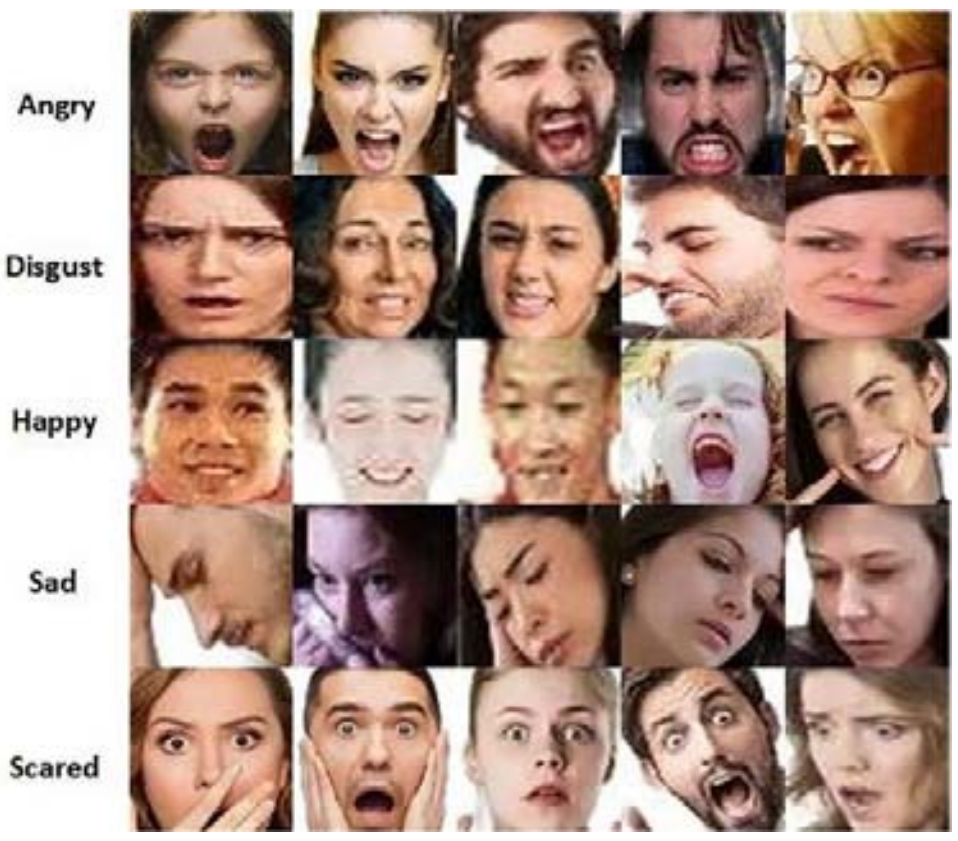

Figure 1 Facial emotion dataset

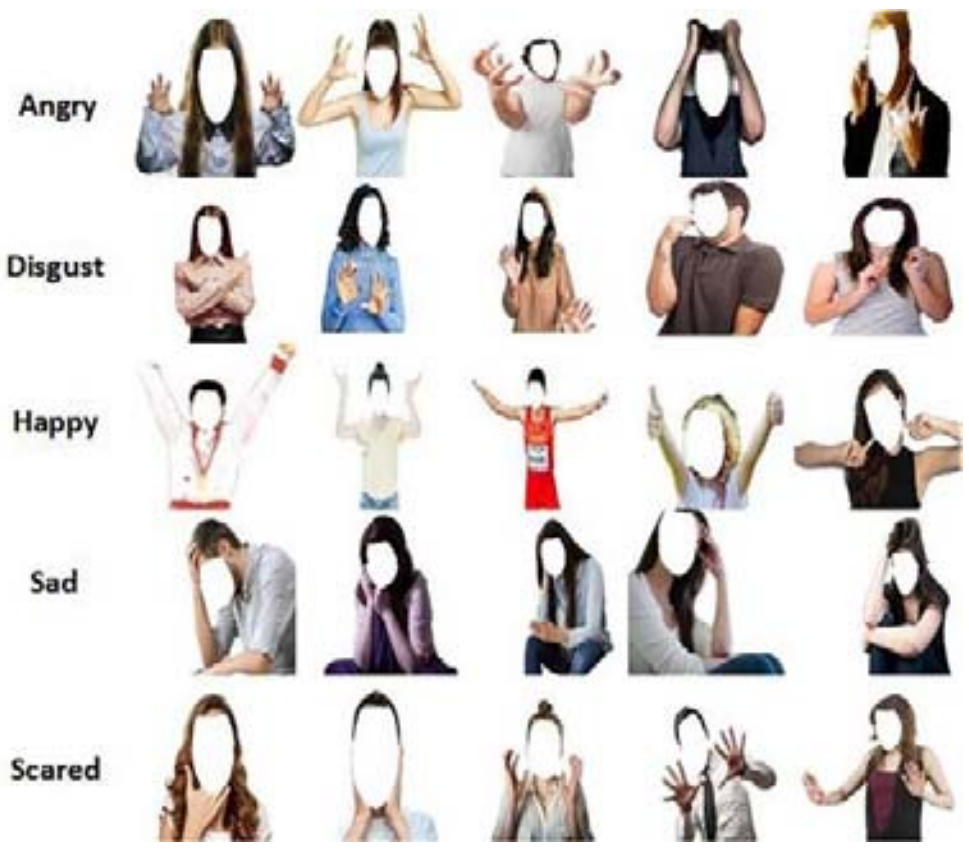

Figure 2 Physical emotion dataset

Table 1 Number of samples in the dataset

| | Angry | Disgust | Happy | Sad | Scared | Total |
|---|---|---|---|---|---|---|
| Image | 3359 | 4287 | 4369 | 2706 | 2720 | 17441 |
| Text | 3359 | 4287 | 4369 | 2706 | 2720 | 17441 |

## 4. Approach

### 4.1 framework and structure

The flowchart of the method in this paper is shown in Figure 3. For the portrait data, it is firstly pre-processed to separate facial expressions and body movements; then facial, body and text features are extracted using Resdual network and Bidirectional Encoder Representations from Transformers respectively; finally, after feature-level fusion to train the fused features, the results of emotion recognition are obtained.

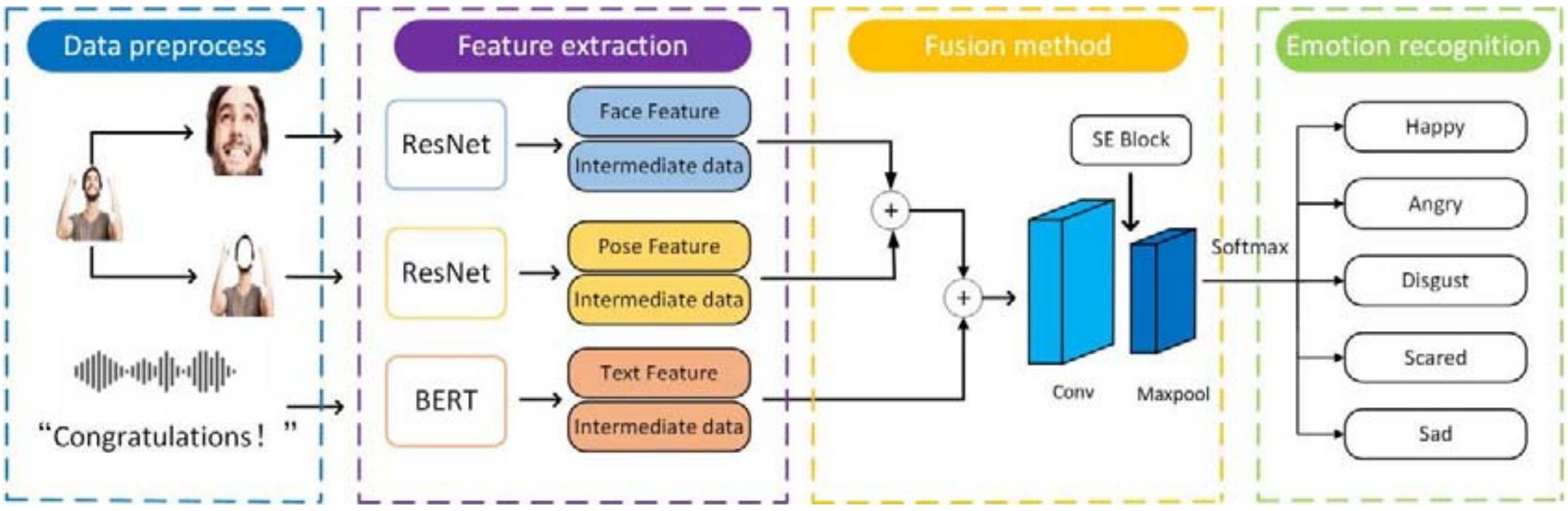

Figure 3 The general flow chart of the method in this paper

**4.2 Data pre-processing**

The image dataset is segmented into face and body by Opencv-DNN and Dilb-Face, and the size was normalized to 224*224. The purpose of using methods is to complement each other and reduce missed detections, and it is not easy to show the full body movements of the person when segmenting the body with Opencv-DNN, so Dilb-Face is used to detect the 68 key points of the face [26] before segmentation.

**4.3 Feature input layer**

The feature input layer consists of feature data from three modalities, namely face data, body data and text data. Each of the three-modal data was extracted from the primary representation by a linear network and the feature data was aligned, i.e., the three modal data are two-dimensional vectors with the same length in the first dimension but not in the second dimension. The text features are BERT word vectors with dimension 768; the facial data and body data were obtained as feature input vectors through the ResNet network framework, both with dimension 2048.

The feature information of the three modalities is shown as follows.

$$N = \{F_f, F_b, F_t\} \quad (1)$$

$$X_n^i = [x_n^i : i \le I,\ x_n^i \in R^{d_{x_n}}, n = N] (2)$$

where is the set of $F_f$ (Face), $F_b$ (Body) and $F_t$ (Text), $n$ containing the information of the three modalities, $i$ is the sample sequence, $I$ is the total sample length, $d_{x_n}$ is the size of the modal feature dimension, $X_n$ is the matrix feature representation of the modality.

**4.4 Feature fusion**

The facial and body feature data were decomposed from the portrait image, and both have the same number of channels and similar semantics of the feature map, so the first fusion was performed. The fused feature vector was then fused with the text features for a second time to obtain the combined features, which are represented as follows.

$$Z(f,b,t) = \sum_{i=1}^{I} (X_{F_f}^{i} \oplus X_{F_b}^{i} \oplus X_{F_t}^{i}) \quad (3)$$

The combined features were further mined for high-level complementary information between multimodal information using convolutional layers, and SE blocks were added to assign different weights to the features.SE blocks act as an attention mechanism on the feature graph to motivate important features and suppress unimportant ones [27]. The specific representation is as follows.

$$\mathrm{Cov}(f,b,t) = \sum_{m=1}^{c} Conv2D(Z^{i}(f,b,t), K_c) \quad (4)$$

$$z_c = F_{sq}(Cov(f,b,t)) = \frac{1}{W \times H} \sum_{i=1}^{W} \sum_{j=1}^{H} H_c(i,j) \quad (5)$$

$$s = F_{ex}(z,W) = \sigma(g(z,W)) = \sigma(g(W_2 \delta(W_1 z)))$$
$$W_1 \in \mathbb{R}^{\frac{C}{r} \times C},\ W_2 \in \mathbb{R}^{C \times \frac{C}{r}},\ c \in \{1,2,...,c\} \quad (6)$$

$$\tilde{X}_c = F_{scale}(Cov(f,b,t), s_c) = s_c \cdot Cov(f,b,t) \quad (7)$$

where c represents the number of channels, Conv2D represents the two-dimensional convolution function, K represents the convolution kernel, $\sigma$ represents the sigmoid activation function, $\delta$ represents the relu activation function, $W_1 \in \mathbb{R}^{\frac{C}{r} \times C}$, $W_1 \in \mathbb{R}^{C \times \frac{C}{r}}$ are the weight matrices of the two fully connected layers, respectively.

The processed combined features were retained the strongest feature information by maximum pooling, and the final classification prediction results were output through the fully connected layer as follows.

$$\tilde{X}_C = \max pooling\{\tilde{X}_c\} \quad (8)$$

$$Y^{(i)} = \partial(W_C \cdot \tilde{X}_C^{(i)} + b_C) \quad (9)$$

where i is the index of the combined features, $\partial$ is the Softmax activation function, $W_C$ and $b_C$ are the weights and biases of the Softmax layers.

### 4.5 Optimization strategy

Cross entropy was used as the loss function in all model training processes, and the formula as follows

$$Loss(Y, y) = -\sum_{i=1}^{I}\sum_{c=1}^{C} y_i^c \cdot \log Y_i^c \quad (10)$$

Where, y denotes the true label, Y denotes the predicted label, I denotes the total number of training samples, and C denotes the number of categories. The Adam (Adaptive Moment Estimation [28]) optimizer was also used to optimize the parameters of the network.

---

**Algorithm 1** Feature after Feature

---

**Data**：B：batches of training data

$B = \{b_1, b_2, ...b_n\}$

**Input**: Face feature: $F_f$ Body feature: $F_b$ Text feature: $F_t$

**Output :** Fusion feature vectors: $\tilde{X}_C$

Function *feature after feature*

If $F_f$ != NULL, $F_b$ != NULL, $F_b$ != NULL:

Logitscale ← Self-learing weighting parameters

for each $e_i \in F_f$ $e_j \in F_b$ $e_k \in F_t$ do

if *loss* Continuing decline then

$E_{fusion} \leftarrow e_i \oplus e_j \oplus e_k$

else

$E_{fusion} \leftarrow Logit_{scale} \cdot (e_i \oplus e_j \oplus e_k)$

End if

$Cov_{fusion} \leftarrow Conv2D(E_{fusion})$

$z_{sq-fusion} \leftarrow F_{sq}(Cov_{fusion})$

$s \leftarrow F_{ex}(z_{sq-fusion}, W)$

$\tilde{X}_c \leftarrow s \cdot Cov_{fusion}$

return $\tilde{X}_c$

end for

end if

end fuction

---

## 5. Experiments and Results

### 5.1 Experimental setup and environment configuration

The operating system environment used in this paper is Windows 10; the CPU version is Intel(R) Xeon(R) Gold 5215M; the GPU version is GTX 1080Ti; the programming language is Python 3.9; the deep learning environment is PyTorch 1.10.

### 5.2 Evaluation metrics

$$\mathrm{Re}call = \frac{TP}{TP + FN}$$

$$\mathrm{Pr}ecision = \frac{TP}{TP + FP}$$

$$F1 - score = \frac{2 \times TP}{2 \times TP + FN + FP}$$

$$Accuracy = \frac{TP + TN}{TP + TN + FP + FN}$$

In this paper, Recall, Precision, F1-score, Accuracy, and confusion matrix and ROC curve are mainly used as the evaluation metrics of the model. The experimental metrics were calculated as follows.

Where, TP denotes the number of actual true categories and predicted as true categories; TN denotes the number of actual wrong categories and predicted as wrong categories; FN denotes the number of actual wrong categories and predicted as true categories; FN denotes the number of actual true categories and predicted as wrong categories.

### 5.3 Results Analysis

#### 5.3.1 Unimodal sentiment recognition experiments

The unimodal emotion recognition experiments were conducted for facial expressions, body gestures and text data. VGG16 [30], ResNet50 and ViT [31] were selected for comparison to determine the network for feature extraction in multimodal fusion for the image dataset, and LSTM [32] and BERT were selected for the text dataset to determine their feature word vectors. Figure 4 indicates the confusion matrix of each network under single modality, and Figure 5 indicates the ROC curves of each network under single modality.

Table 2 Precision, recall, F1 value and correctness of different networks under facial, posture and text emotion

| | Modal | Precision | Recall | F1 | Accuracy |
|---|---|---|---|---|---|
| Facial | VGG16 | 0.788 | 0.7742 | 0.7765 | 0.7881 |
| | ResNet50 | 0.8132 | 0.8041 | 0.8079 | 0.8192 |
| | ViT | 0.6752 | 0.6708 | 0.6688 | 0.6757 |
| Body | VGG16 | 0.6949 | 0.6835 | 0.6832 | 0.693 |
| | ResNet50 | 0.7515 | 0.7322 | 0.7359 | 0.7437 |
| | ViT | 0.4925 | 0.4948 | 0.479 | 0.4977 |
| Text | LSTM | 0.4578 | 0.4142 | 0.4035 | 0.3877 |
| | BERT | 0.6274 | 0.6205 | 0.6225 | 0.6213 |

Table 2 Results for facial emotion, body pose and text in different neural network models.

The facial and body modality results show that ResNet50 performs best compared to the traditional VGG16 for the same picture, which is inseparable from its unique residual connectivity and deeper convolutional layers. The use of ViT as a transformer in vision is slightly less effective, which may be related to the size of the dataset and the complexity of the expressions it faces. In the natural language domain, however, the transformer-based BERT network outperforms the LSTM by far, thanks of course to its special attention mechanism.

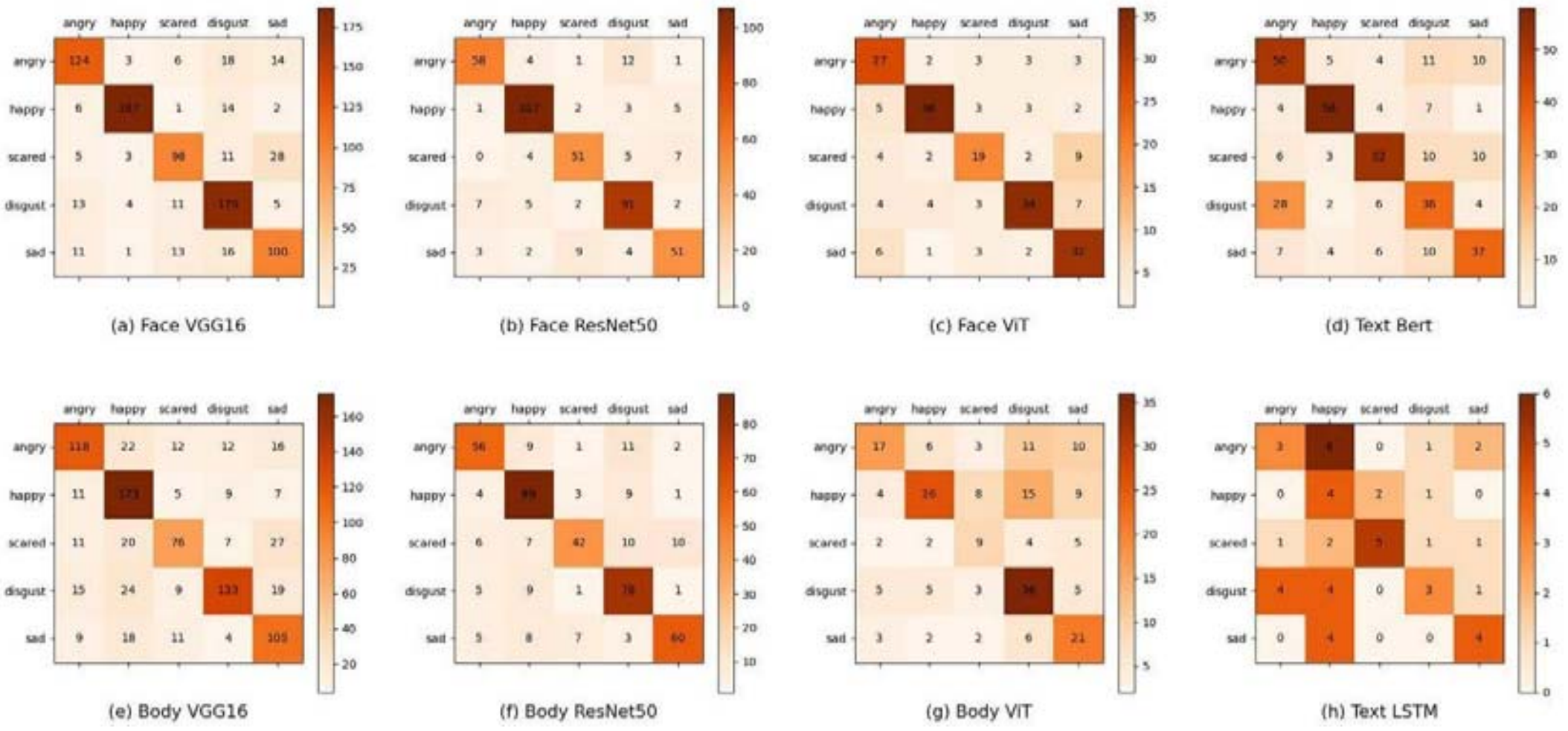

Figure 4 Confusion matrix for each network in unimodal mode, (a) (b) (c) (d) (e) (f) (g) and (h) correspond to the confusion matrix of dimensions Face VGG16, Face ResNet50, Face ViT, Text Bert, Body VGG16, Body ResNet50, Body ViT and Text LSTM respectively.

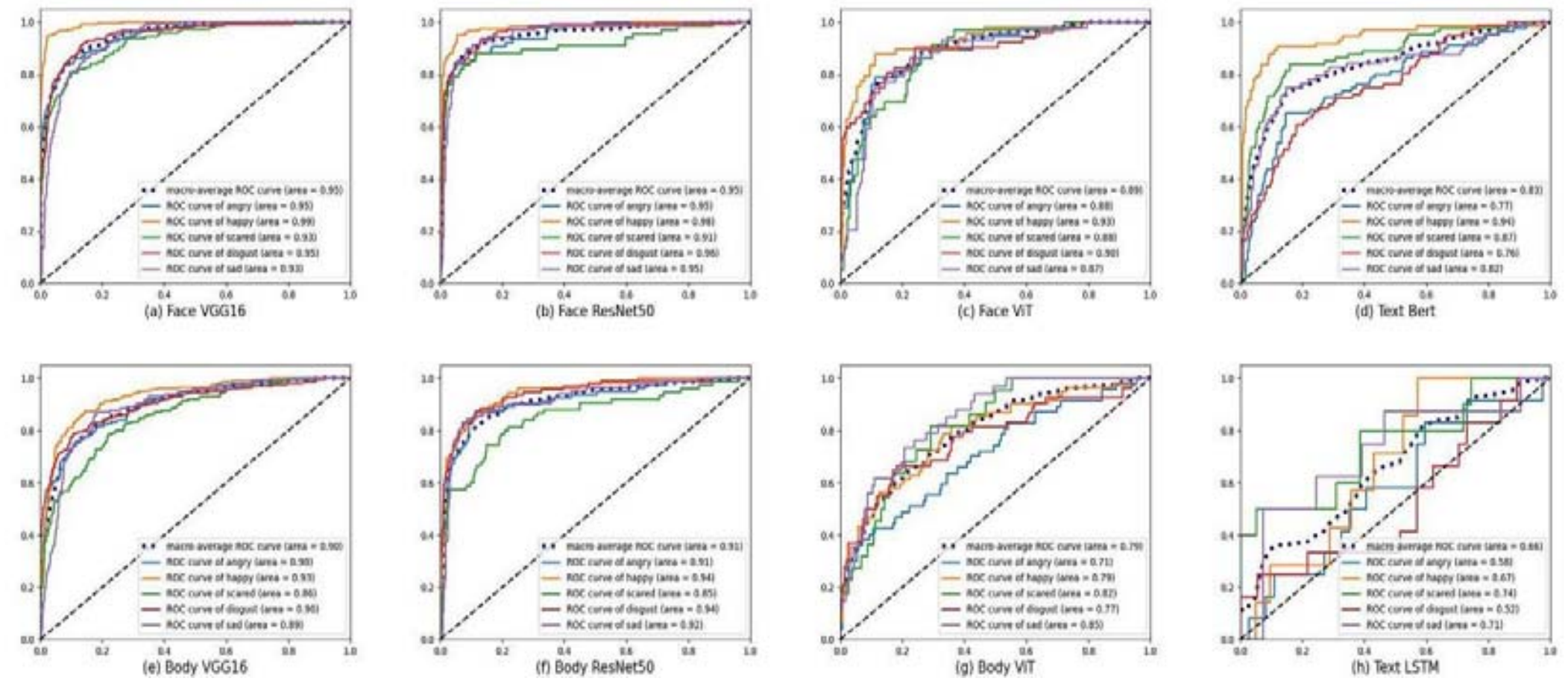

Figure 5 ROC curves for each network in unimodal mode, (a) (b) (c) (d) (e) (f) (g) and (h) correspond to the confusion matrix of dimensions Face VGG16, Face ResNet50, Face ViT, Text Bert, Body VGG16, Body ResNet50, Body ViT and Text LSTM respectively.

The ROC curve results in Figure 5 illustrate that happy and disgust have the highest correct rates among the five emotion classes, while scared has the lowest correct rate of recognition. These results may be related to the recognition characteristics of emotion, and the performance of the algorithm model. For example, when people feel happy, they may uplift corners of the mouth, dimples, and with the corners of the mouth dropping while disgust. Besides, scared has a variety of expressions related to personal habits, therefore has the worst recognition results. The rest of the emotions have fewer outwardly expressed features, and it is difficult for the classifier to fully learn the corresponding emotion.

### 5.3.2 Multimodal sentiment fusion experiments

The unimodal feature vectors have been obtained in section 4.4.1 by emotion recognition of face, body and text, respectively. ResNet50 network was chosen for both face and body; BERT network was chosen for text. Here multimodal feature fusion experiments will be performed. The unimodal feature vectors were fused two by two as well as the final fusion of three feature vectors. To demonstrate the model generalization capability, this paper transitions the five-classification problem in a generalized environment to a classification problem in a specific scenario such as a sports environment. Sports events are mostly positive and negative, so this paper designates sports events as a binary classification problem. We compare the proposed method with the methods shown in Table 4, where SOTA1 and SOTA2 denote the current models with the best performance and the second-best performance, respectively. Table 3 shows the results of corresponding metrics after multimodal fusion; Figure 6 shows the confusion matrices of two-two fusion and last three-mode fusion for each modality respectively; Figure 7 shows the ROC curves of two-two fusion and last three-mode fusion for each modality respectively; Figures 8, 9, 10 show the original sample dispersion, sample normalized dispersion, three- The dispersion of the original sample, the dispersion of the sample after normalization, and the dispersion of the cluster after trimodal fusion are shown in Figures 8, 9, and 10, respectively. Figure 11 show the Grad-CAM[35] attention mechanism result of computer and human.

Table 3 Precision, recall, F1 value and Accuracy after multimodal fusion in common and sport scence

| | | Precision | Recall | F1 | Accuracy |
|---|---|---|---|---|---|
| Common scence | Face+Body | 0.7787 | 0.7792 | 0.7676 | 0.7734 |
| | Face+Text | 0.7078 | 0.6969 | 0.6845 | 0.7262 |
| | Body+Text | 0.6329 | 0.6276 | 0.6277 | 0.6567 |
| | Face+Body+Text | 0.8195 | 0.8196 | 0.8190 | 0.8375 |
| Sport scence | Face | 0.9736 | 0.9285 | 0.9480 | 0.9611 |
| | Body | 0.9218 | 0.9642 | 0.9391 | 0.9484 |
| | Face+Body | 0.9666 | 0.9791 | 0.9721 | 0.9736 |

Table 4 Comparison of FAF with current best methods

| | Precision | Recall | F1 | Accuracy |
|---|---|---|---|---|
| SVM | 0.6689 | 0.6541 | 0.6569 | 0.6706 |
| CentralNet[33] | 0.8053 | 0.8047 | 0.8024 | 0.8215 |
| LWF[34] | 0.7326 | 0.6963 | 0.6964 | 0.7214 |
| FAF | 0.8195 | 0.8196 | 0.8190 | 0.8375 |

Comparison With current methods: Tables 7 show the comparison results of FAF with current best methods. We can clearly see that FAF is outperform other methods. This allows us to conclude that FAF is a good choice when dealing with complex and imbalanced images. Unsurprisingly, standard SVM performs worst of all of the evaluated algorithms, while three other methods try to offset their inability to handle spatial properties of data with advanced instance generation modules. Both CentralNet return the best results from all four tested algorithms, with LWF coming close to GAN-based methods. This can be attributed to their compound oversampling solutions, which analyze the difficulty of instances and optimize the placement of new instances, while cleaning overlapping areas. However, this comes at the cost of very high computational complexity and challenging parameter tuning. FAF returns superior balanced training sets compared to pixel-based approaches, while providing an intuitive and easy to tune architecture and, according to both nonparametric and Bayesian tests presented in Table IV, outperforms all pixel-based approaches in a statistically significant manner.

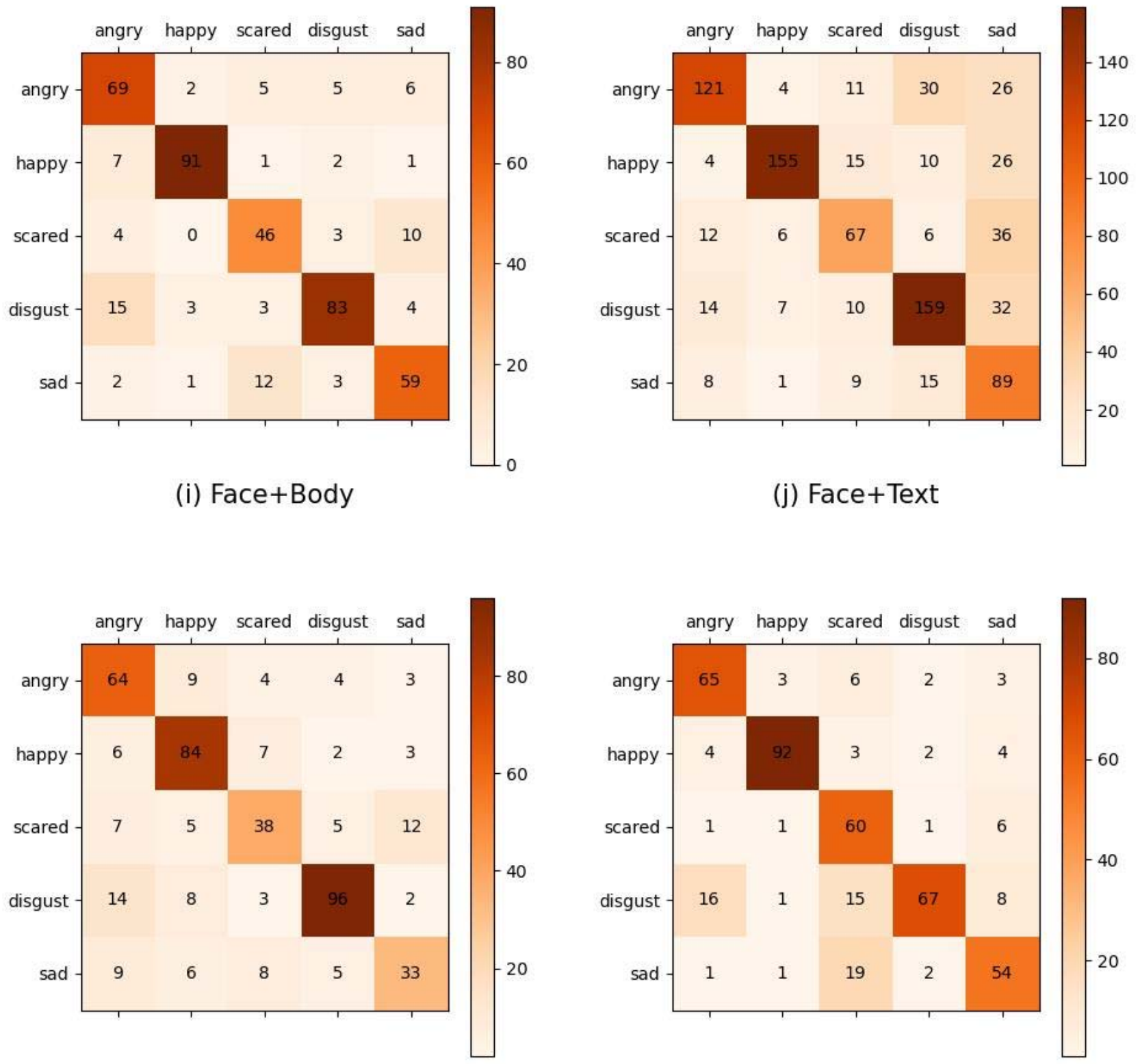

(i) Face+Body

(j) Face+Text

(k) Body+Text

(l) Face+Body+Text

Figure 6 Multimodal fusion confusion matrix, (i) (j) (k) and (l) correspond to the confusion matrix of dimensions Face+Body, Face+Text, Body+Text, Face+Body+Text and Text LSTM respectively.

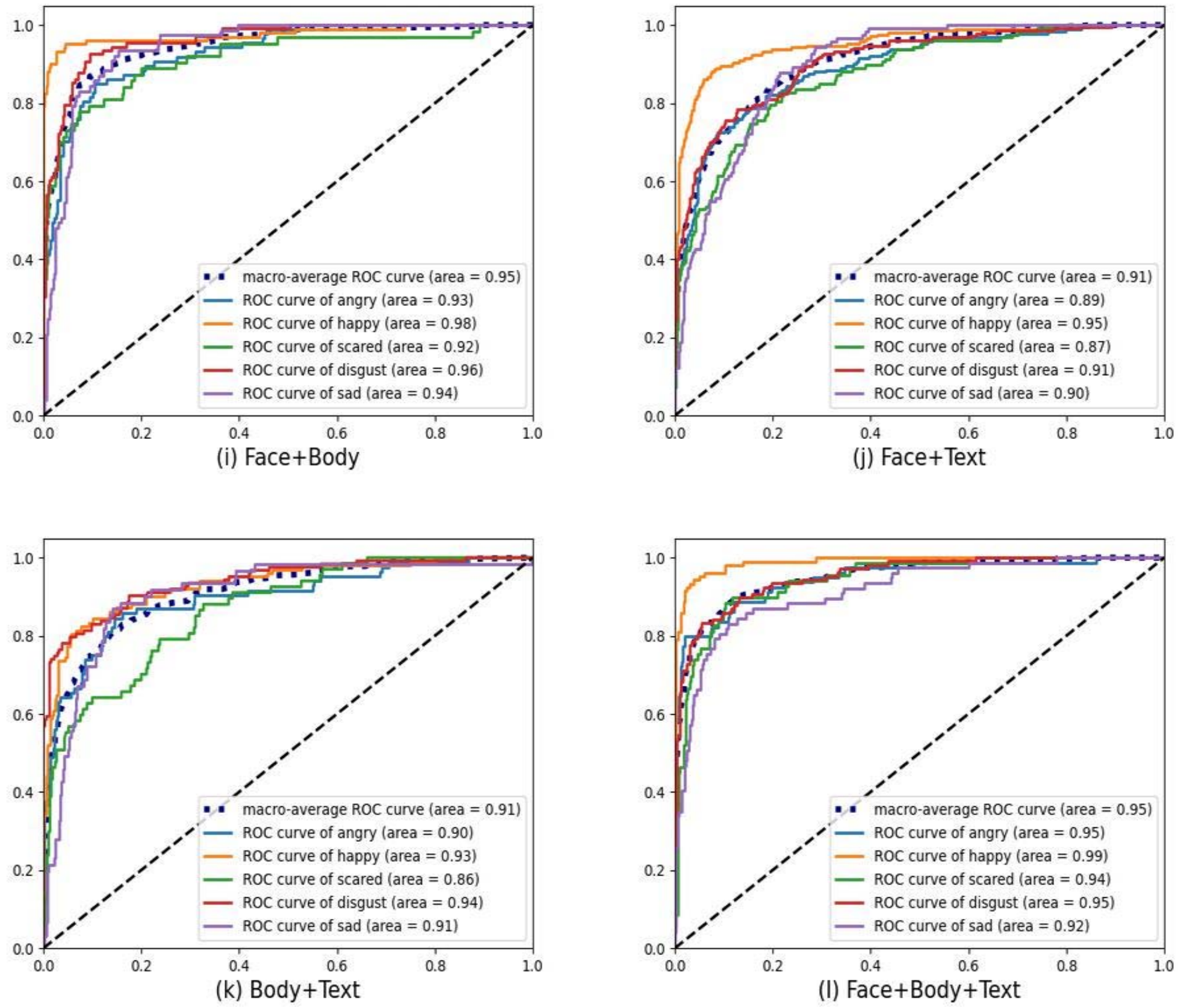

Figure 7 Multimodal fusion ROC curve, (i) (j) (k) and (l) correspond to the confusion matrix of dimensions Face+Body, Face+Text, Body+Text, Face+Body+Text and Text LSTM respectively.

Figures 6 and 7 show comparison results between the bimodal and trimodal. It can be seen that happy and disgust perform best in the bimodal case, which may be related to the datasets; scared performs the worst, which may be due to the reaction time factor associated with fear, just as when we show fear, the pupils may first dilate and be accompanied by This may be due to the reaction time factor associated with fear, as when we show fear, the pupils may first dilate and be accompanied by a series of coherent responses such as mouth opening and postural changes. In the trimodal case, however, this situation improves, as the network is no longer biased towards one emotion, and the five emotion scores are more evenly distributed. This is also supported by the ROC curves in Figure 7, where happy and disgust show much lower error rates than scared, and are more balanced in the trimodal case.

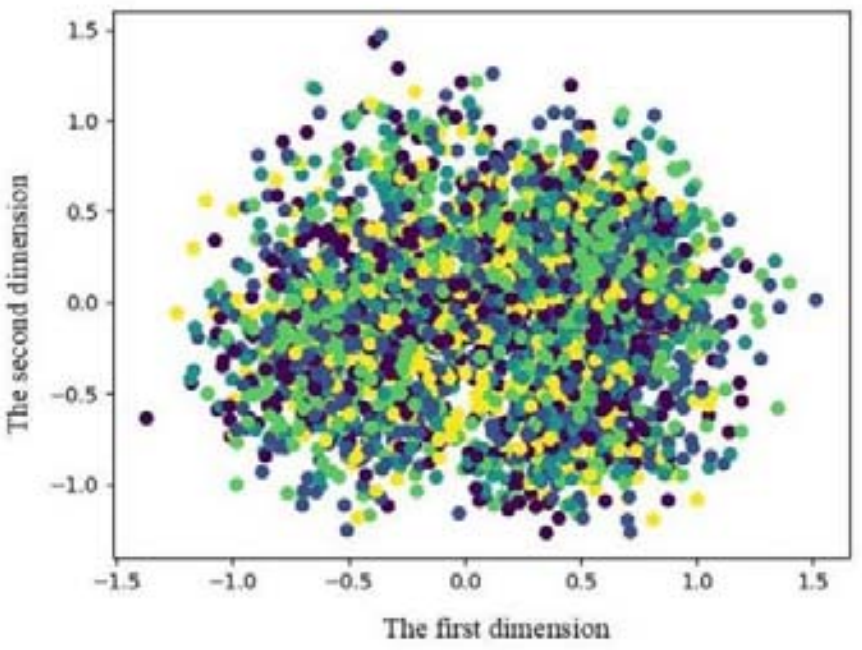

Figure 8 Original sample dispersion

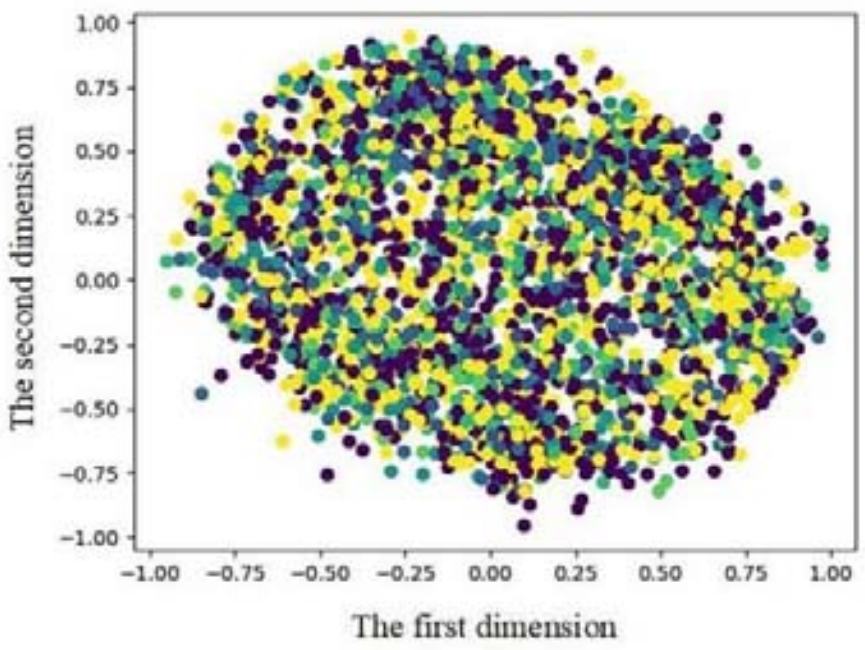

Figure 9 Sample normalization

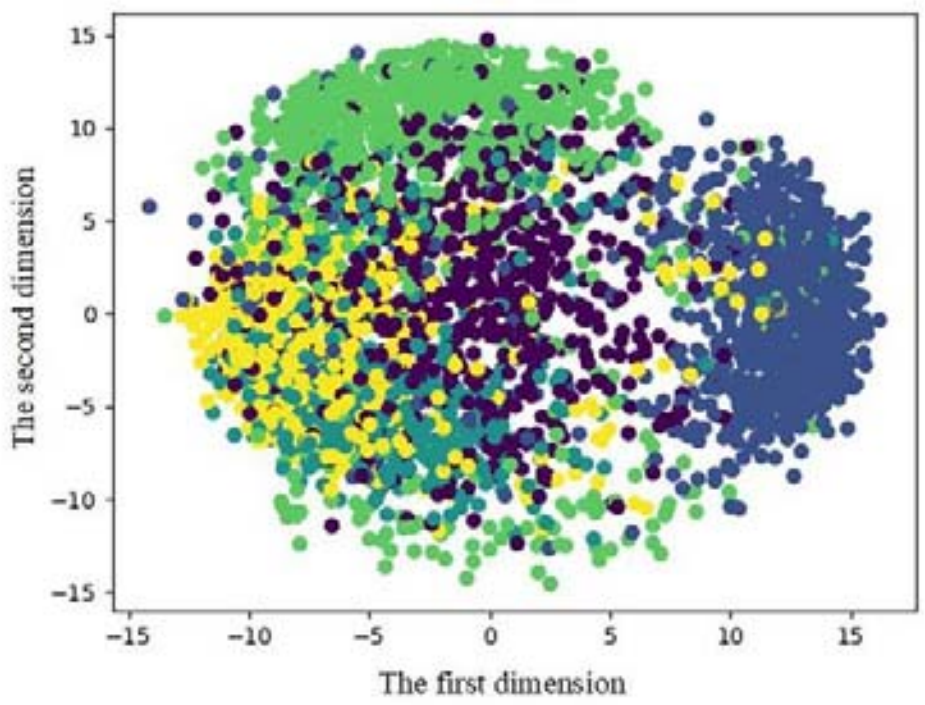

Figure 10 Sample dispersion after fusion clustering

Figures 8, 9 and 10 show the clustering effect of the HED dataset when it is subjected to the FAF framework. We can clearly see that the initial dataset is randomly disrupted when entering the network, then it is normalised by a normalisation operation and finally classified into five sentiment categories by the FAF classifier. As the image shows, most of the five sentiments can be accurately classified.

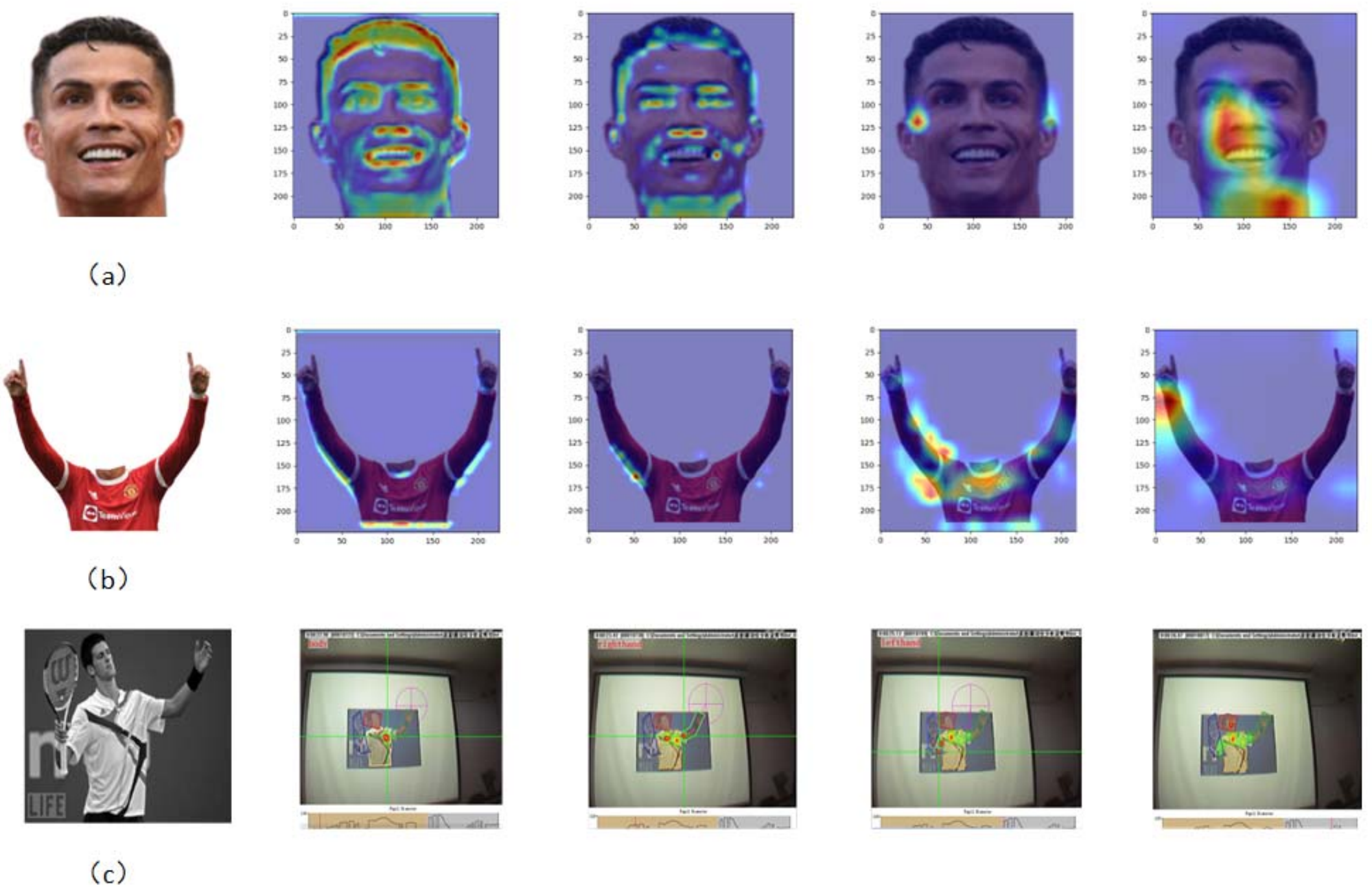

Figure 11 attention mechanism result of computer and human

The experimental results show that the recognition rate of each modality fusion exceeds the recognition rate of the corresponding single modality under the premise of using the method of this paper, which illustrates the effectiveness of fusing facial expressions, body movements and text for emotion recognition. Comparing the recognition effects of modal fusion, it can be seen that the combination of features with high-level semantic information extraction and attention mechanism can effectively alleviate the problems of information redundancy and high dimensionality, and improve the accuracy of emotion recognition in single modality by about 1.83%~21.62%.

We also compared the responses of Resnet and human eye-movement metrics in the face of different emotional expressions. Figures (a) and (b) represent the results of the attention visualization of facial emotions and body movements when passing through the FAF framework, respectively. The first image in figure (a) is the original image, while the second, third, fourth and fifth images show the results of passing through the first, second, third and fourth convolution blocks, respectively. It can be seen that FAF is more accurate in capturing the features of each modality. In the figure, when the character shows a pleasant emotion, the corresponding features are more concentrated on the teeth, dimples, etc. Similarly, to the face, when the character shows pleasure, the physical features tend to be more on the upraised arm and finger posture. Figure (c) shows that the human eye visualizes attention through the oculomotor when judging the emotion of a person. As can be seen, the human eye observes the limbs first, followed by the right hand, then the left hand, and finally the face. However, the neural network clearly outperforms the oculomotor in terms of more granular and fine-grained emotion judgements. Operational time plus, respective advantages and disadvantages

## 6 Online multimodal emotion prediction platform

Considering the verification of the accuracy of multimodal emotion fusion experiments and the significance of future applications, several models trained during the experiments were saved and a multimodal emotion online prediction platform was developed. The platform provides four emotion testing portals, namely, unimodal facial expression prediction, unimodal body movement prediction, unimodal text prediction, and arbitrary multimodal fusion prediction(see Figure 12).

### 6.1 System Architecture

The platform was designed with a front-end and back-end separated architecture model. The front-end part was developed based on the progressive JavaScript framework Vue.js, and combined with Element UI library, Echarts chart library, Axios network request library, etc. The operation is simple and the interface is intuitive, with good user experience; the back-end part is based on the Python Django web application framework development, using Django Rest Framework interface design style, and deployment of PyTorch deep learning model.

When the user enters the operation page and selects an image file or enters text content and clicks submit, the browser declares a FormData object to simulate the form submission data, the image file is converted into a binary data stream and uploaded asynchronously, and the text content is parsed into string type data and sent to the specified server interface in the form of key-value The text content is parsed and sent to the specified server-side interface in the form of key-value pairs using the POST request method of the HTTP protocol.

After receiving the data from the user, the server-side interface first preprocesses the image according to the user's selection in the browser, saves the segmented facial or body parts, then loads the corresponding PyTorch model, inputs the processed data into the model one by one in the evaluation mode, and obtains the scores of the five emotion categories output by the model, and returns the information on the key points of the face collected in the data preprocessing process as JSON data. The information is returned to the client as JSON data type, and finally the client calls the relevant library to visualize and report the data returned from the server.

### 6.2 Model saving and deployment

The server-side PyTorch 1.11.0 was installed on top of the Django 3.0 web development environment as the environment for model loading and evaluation. The experimenter serializes and saves the model parameters to the server disk after training a specified number of rounds. When it is necessary to input data into the model to get evaluation results, a model object is first initialized, and then the model parameters were read and loaded from the disk. Seven PyTorch models were deployed on the server side, corresponding to unimodal evaluation of facial expressions, body movements, and verbal text, bimodal fusion evaluation of face and body, face and text, and trimodal fusion evaluation of face, body, and text, respectively.

Considering the problem that the server side repeatedly loads models every time the user side requests an interface, which wastes resources and has low performance, the developers write the code for loading models into the ready() method of Django's AppConfig class, so that no matter how many times the user side requests an interface, all models were only loaded automatically when the server side project starts, and the relevant interfaces directly call the pre-loaded The relevant interface directly calls the pre-loaded model for evaluation, effectively improving server-side performance and evaluation efficiency.

## 6.2 Future Prospects

At present, the multimodal emotion online prediction platform is developed for relevant experimental researchers, invoking deep learning models to evaluate samples in real time and realize data reports. In the future, based on further expansion of the data set and improvement of model prediction accuracy, more emotion-related functions will be combined and put online to provide free emotion prediction for all users.

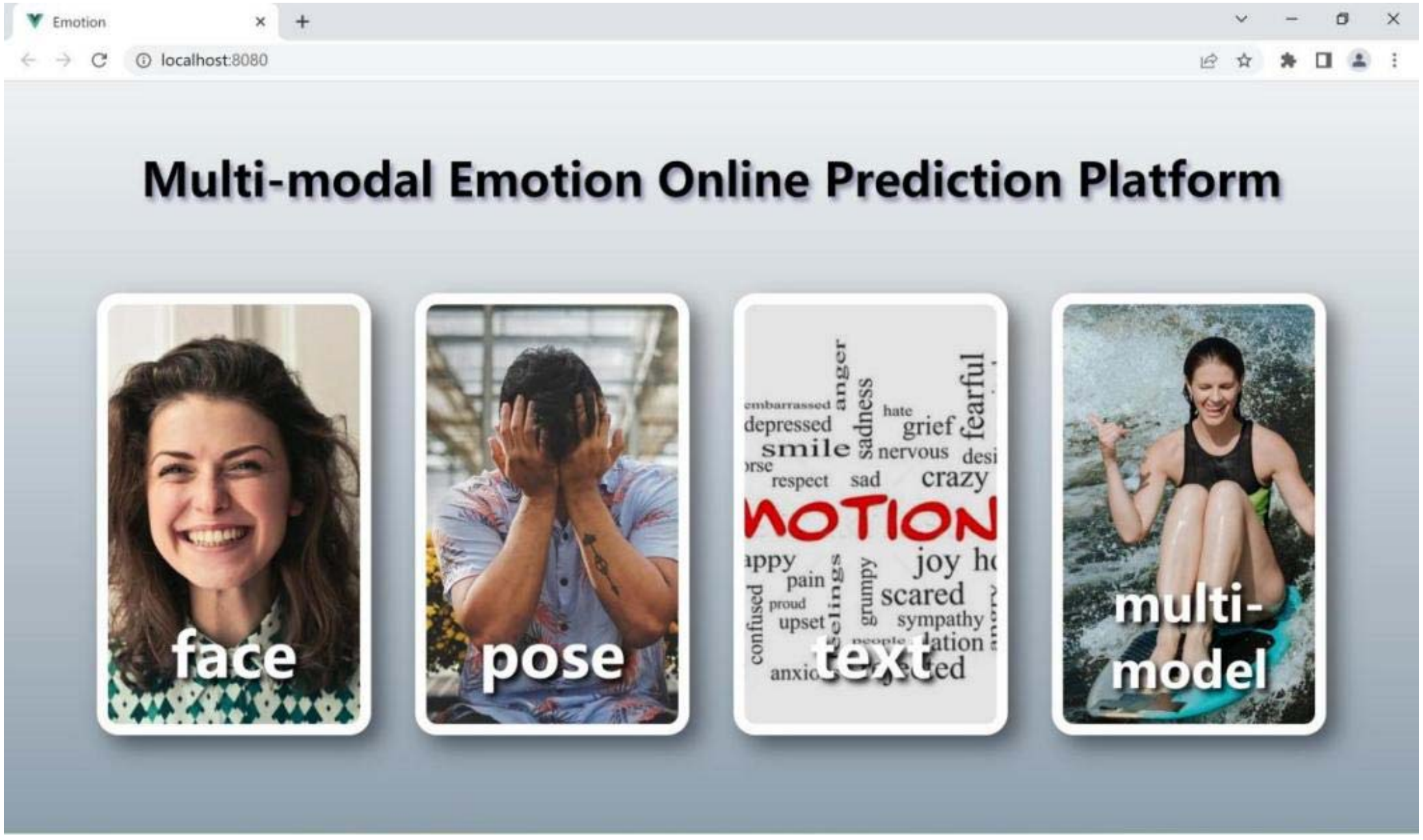

(a) Web page

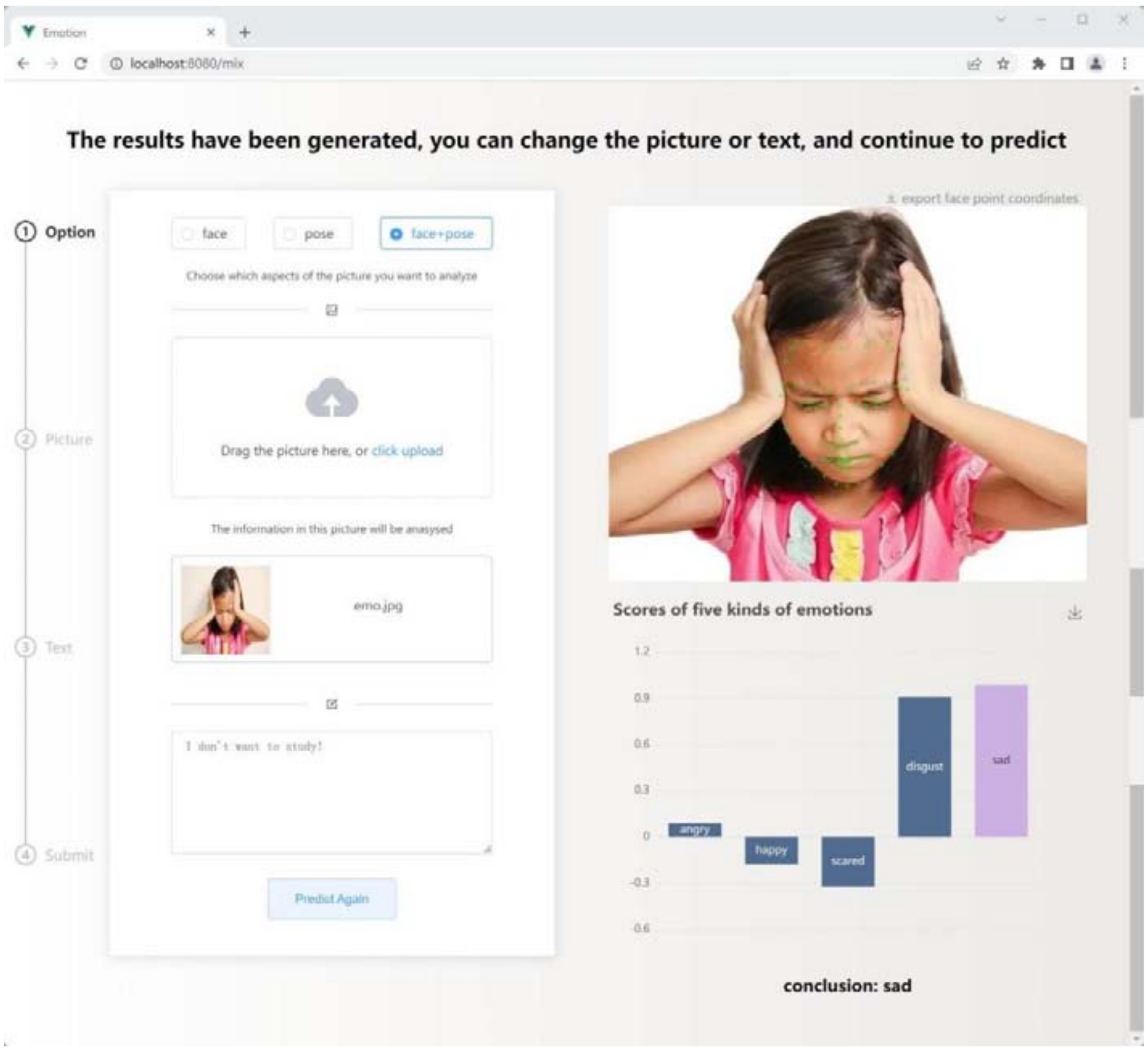

(b) User Interface

Fig 12 Online multimodal emotion prediction platform

## 7 Key Findings

Multimodal emotion recognition outperforms unimodal emotion recognition in most cases, allowing for the fusion of the main features that each unimodal mode. In this study, it was found that the unimodal modality showed better recognition accuracy for the significant positive and negative emotions in the HED dataset, such as happy and disgust. The Face, Body and Text modalities were investigated separately and found to have the highest recognition accuracy for the Face modality, followed by Body and Text modalities in the HED dataset. On the other hand, the ResNet network structure and the BERT network structure both showed the highest Precision, Recall, F1 and Accuracy for the five sentiment categories in unimodal sentiment recognition. Transformer-based strong feature extraction capability.

In the multimodal state, the trimodal model outperformed all the bimodal models. Some of the bimodal models did not improve on their unimodal counterparts, probably because the remaining modalities were less relevant in the absence of major emotion expression features, such as facial emotion. In the fusion operations associated with the facial emotion modality, the results were all optimal. Also similar to unimodality, strong positive and negative emotions showed the highest recognition rates in bimodal recognition. However, the performance of each emotion decreases for fusion operations lacking the facial emotion modality. However, in the trimodal fusion operation, the recognition of each emotion was more integrated and no longer biased towards strong positive and negative emotions.

Context is also a factor that influences sentiment analysis. In order to explore the problem of emotion recognition in specific scenarios, the context of the emotion is defined as a sporting event, and the five-category problem in the generalized state is replaced by a two-category problem with strong positive and negative emotions. The analysis shows that the recognition rates of facial expressions and body gestures do not differ much and the fused recognition error rate is lower, provided that the background is determined. After removing the background problem and the remaining compound expressions from the five classifications, body movements contributed to the recognition of athletes' emotions, which is consistent with Ahmed F et al.'s study [36].

## 8 Implications, Limitations and Future Research

Taken together, we developed a real-world multimodal dataset "HED" that expanded the existing multimodal dataset and FAF, a deep learning algorithm based on image and text information, was used for emotion recognition. More importantly, our findings highlight significant variances based on important unifacial factors, like posture and the context information. Notably, our findings add to a growing stream of literature that cautions about the prevalence of biases in machine learning datasets and models [37]. Our findings highlight that the relative performance of emotion recognition systems might vary substantially across gesture and context. Researchers and developers need to be aware of such variances, especially if such systems are being used in a gesture-sensitive context.

our study supports previous research in human emotion recognition. Multimodal based method could improve the emotion classification efficiency. Recently, the face expression, gesture and the general background data from the scene are considered as the complementary cues for emotion prediction. However, most of the existing works still have some limitations in deeply exploring the scene-level context when it is not clear enough. So, we try to extract text information from the emotional state prediction method based on visual relationship detection between human and the text from the background. After that, the model incorporates those features with context and body features of the target person to predict their emotional states.

There are some limitations in this study that need future work. First, the current study developed a training and test dataset based on face, gesture and text as the three data modalities. A multimodal emotion recognition method was used to compare with others. Not surprisingly, although the latest and more effective cross-fusion model was used, the quality of the dataset is still key to limiting the recognition efficiency. Different people have different perspectives on emotional expression and it may influence the development of emotion dataset. This may change the fusion feature, leading to a significant different classification result. Second, while our dataset provided good performance in emotion classification, as we noted, the HED dataset, though with a relatively big data size, but included relatively fewer representations of disgust, fear and surprise which prevented us from promoting the classification efficiency. The annotators' demographic distribution was not analyzed, which prevented us from studying the sensitivity of our findings.

Last, in this study, three important factors: face, posture and text were selected for emotion classification. Other factors like scene background, physiological contexts may also affect the recognition performance. However, these factors may fusion into context feature which sometime use text feature instead. Background provided key evidence when difficult to determine the emotional state, which was not sensitive to the real-time emotion recognition. Further research is needed for deep semantic feature mining and emotion recognition in dynamic environment. Future work in emotion recognition needs high quality dataset, recognition models and more advanced theoretical models to interpretation of multimodal emotion recognition. This not only increase the efficiency of sentiment recognition, but also helps to improve the robustness of the system and provide fine-grained interpretation of the results.

## 8. Conclusion

In this paper, a large "HED" dataset was developed to facilitate the emotion recognition task. To promote recognition accuracy, "Feature After Feature" framework was used to explore crucial emotional information from the aligned face-body-text samples. First, feature extraction was performed for each of the three modalities, and the facial and body features were fused, and then the text feature was fused again into facial-body gesture. The high semantic information of the combined features was then extracted after a two-dimensional convolutional layer and an attention mechanism was added to enhance the important features and weaken the unimportant ones. The experimental results show that the proposed method performs well on multimodal emotion recognition tasks, and the accuracy rate can reach 83.75% after trimodal fusion. The accuracy, recall and F1 values achieve better results, which verifies the effectiveness of the method in this paper.